\documentclass{bmvc2k}
\usepackage{amsmath}
\usepackage{hyperref}
\usepackage{amsfonts}
\usepackage{mathptmx}
\usepackage{algorithm}
\usepackage{algorithmic}
\usepackage{booktabs}   
\usepackage{widetable}
\usepackage{wrapfig}

\usepackage{subfigure}

\title{Cascading Feature Extraction for \\ Fast Point Cloud Registration}

\addauthor{Yoichiro Hisadome}{hisadome@hal.t.u-tokyo.ac.jp}{1}
\addauthor{Yusuke Matsui}{matsui@hal.t.u-tokyo.ac.jp}{1}

\addinstitution{
 Graduate School of Information Science and Technology\\
 The University of Tokyo\\
 Tokyo, Japan
}

\runninghead{hisadome, matsui}{cascading feature extraction for 3d registration}

\def\etal{\emph{et al}\bmvaOneDot}

\begin{document}

\maketitle

\begin{abstract}
We propose a method for speeding up a 3D point cloud registration through a cascading feature extraction.
The current approach with the highest accuracy is realized by iteratively executing feature extraction and registration using deep features.
However, iterative feature extraction takes time.
Our proposed method significantly reduces the computational cost using cascading shallow layers.
Our idea is to omit redundant computations that do not always contribute to the final accuracy.
The proposed approach is approximately three times faster than the existing methods without a loss of accuracy.
\end{abstract}

\section{Introduction}
\label{sec:intro}
Point cloud registration is the task of aligning two 3D point clouds into a single point cloud. 
The two point clouds to be aligned usually represent the same object but are taken from different angles.
Point cloud registration is a fundamental building block for 3D data processing, such as simultaneous localization and mapping and 3D object reconstruction.
\begin{wrapfigure}[15]{r}[2mm]{0.55\hsize}
    \center
    \vspace{-6mm}
    \includegraphics[keepaspectratio, width=\hsize]{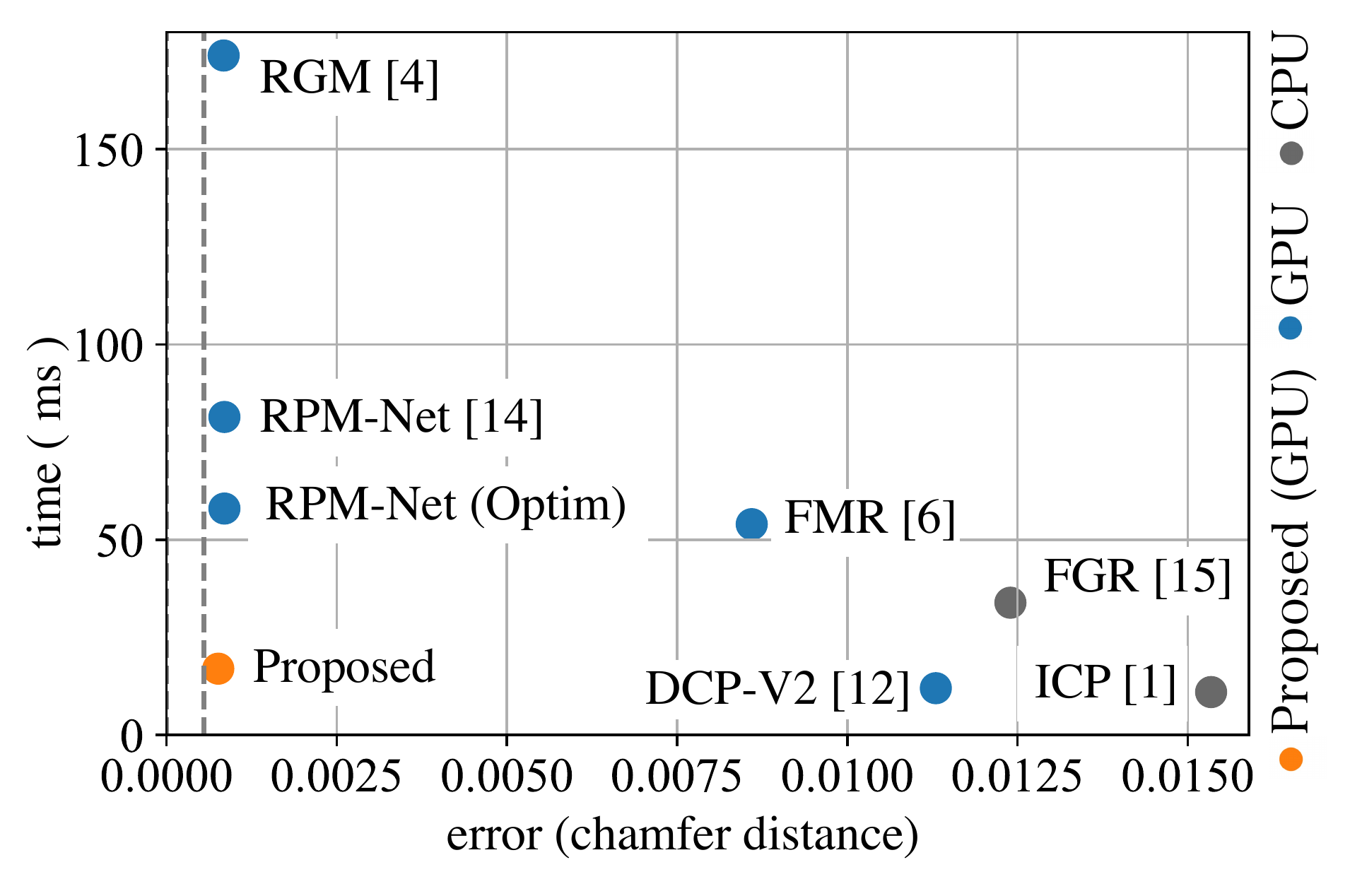}
    \vspace{-8mm}
    \caption{Diagram of accuracy and speed.
        The proposed method achieves both a high accuracy and speed (evaluated on modelnet40~\cite{modelnet40}).
        }
    \label{fig:acc_time}
\end{wrapfigure}

The current state-of-the-art algorithm~\cite{rpmnet} in terms of accuracy takes approximately 60~ms to align approximately 1,000 points on a standard GPU (NVIDIA Tesla V100). 
To achieve a real-time performance, the inference should be less than 30~ms, and less than 20~ms would be even more desirable.

In this paper, we propose a fast 3D point cloud registration method. 
Our idea is to omit the computation of a part of the iterative process that is not necessarily accurate.
Existing methods repeat the same heavy processing during every iteration.
With the proposed method, the heavy processing runs only once initially, and the light processing is repeated in the subsequent iterations. 
Through this approach, we can significantly reduce the overall computational cost.

RPMNet~\cite{rpmnet} and RGM~\cite{Fu2021RGM} are the most accurate methods available today.
RPMNet and RGM achieve the same level of accuracy, whereas RPMNet is more than twice as fast.
Thus, we used RPMNet for our baseline in this study.
The results and a comparison with other methods are shown in Fig.~\ref{fig:acc_time}.
The proposed method is three times faster than RPMNet without any loss of accuracy.
In addition, our method is more accurate than RGM.

\section{Related Work}
\begin{figure}
\includegraphics[width=\textwidth]{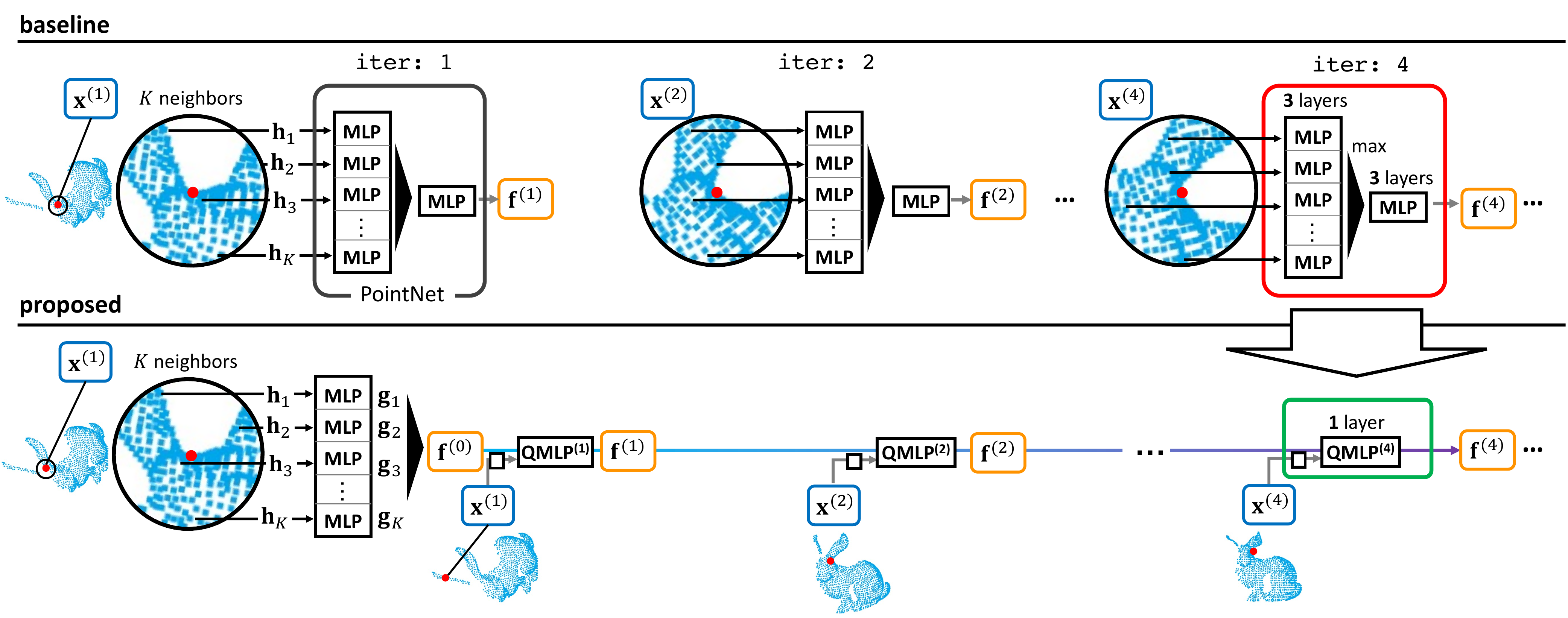}
\vspace{-5mm}
\caption{Overview of our method, with comparison to the baseline RPMNet~\cite{rpmnet}. The subscripts and superscripts denote the index and number of iterations, respectively.}
\label{fig:overview}
\end{figure}
Next, we show a general pipeline of modern point cloud registration methods~\cite{dcp, fgr, rpmnet, Fu2021RGM} and describe in detail the latest approach, RPMNet~\cite{rpmnet}.
Let us assume that there are two point cloud sets, $\mathcal{X} \subset \mathbb{R}^3$ and $\mathcal{Y} \subset \mathbb{R}^3$.
Many modern point cloud registration methods take the following steps.
\begin{enumerate}
    \item Obtain a feature vector for each point, for both $\mathcal{X}$ and $\mathcal{Y}$.
    \item For all combinations of a feature vector in $\mathcal{X}$ and one in $\mathcal{Y}$, compare the distance in a brute force fashion and obtain the correspondence among $\mathcal{X}$ and $\mathcal{Y}$.
    \item Estimate the rigid transformation $T$ that minimizes the sum of the squared distances between the corresponding points.
\end{enumerate}
Several methods, including RPMNet, take the following procedures in addition to (i)$\sim$(iii) above.
\begin{enumerate}
    \setcounter{enumi}{3}
    \item Apply the rigid transformation to one of the objects $\mathcal{Y} \gets T(\mathcal{Y})$ and return to (i). 
\end{enumerate}
Let us describe this in detail for RPMNet.
The upper part of Fig.~\ref{fig:overview} shows the configuration of RPMNet.
First, we describe the feature extraction in (i). Let us consider the extraction of a 96-dimensional feature $\mathbf{f} \in \mathbb{R}^{96}$ from a point $\mathbf{x} \in \mathbb{R}^{3}$ in a point cloud.
We collect the $K$-nearest neighbors of $\mathbf{x}$ and compute a classical feature vector $\{\mathbf{h}_k\}_{k=1}^K$.
Here, $\mathbf{h}_k \in \mathbb{R}^{10}$ is a 10-dimensional feature that consists of (1) the angles between the $k$-th neighbor of $\mathbf{x}$ and $\mathbf{x}$, (2) the absolute coordinate of the $k$-th neighbor of $\mathbf{x}$, and (3) the relative coordinate of the $k$-th neighbor of $\mathbf{x}$ with $\mathbf{x}$ as the coordinate reference.
We obtain a single 96-dimensional feature $\mathbf{f} \in \mathbb{R}^{96}$ by inputting $\{\mathbf{h}_k\}_{k=1}^K$ into PointNet $\mathrm{(PN)}$:
\begin{eqnarray}
    \mathbf{f} = \mathrm{PN}(\{\mathbf{h}_k\}_{k=1}^K).
\end{eqnarray}
PointNet~\cite{pointnet} is a feature extraction module that consists of a multilayer perceptron (MLP) and maxpool.

Next, we describe the computation of the correspondence in (ii).
In the process of (ii) for RPMNet, a soft registration is conducted using simulated annealing.
This makes their method quite robust to noise.
Let $\mathcal{X} = \{\mathbf{x}_i\}_{i=1}^N \subset \mathbb{R}^3 and \mathcal{Y} = \{\mathbf{y}_i\}_{i=1}^M \subset \mathbb{R}^3$ be the two point clouds to be aligned.
Let $d_{i, j}$ be the Euclidean distance between the features of $\mathbf{x}_i$ and $\mathbf{y}_j$.
Herein, we construct a matrix $\mathbf{M} \in \mathbb{R}^{N \times M}$, where the $(i,j)$-th element $m_{i, j} \in \mathbf{M}$ is defined as follows:
\begin{eqnarray}
    m_{i, j} &=& e^{-\beta(d_{i,j}^2 - \alpha)}.
\end{eqnarray}
The parameters $\alpha$ and $\beta$ are positive scalars.
An independent network estimates the optimal values of $\alpha$ and $\beta$.
If $\beta$ is large, $\mathbf{M}$ represents a harder (discrete) correspondence; however, if it is small, $\mathbf{M}$ represents a softer (ambiguous) correspondence.
In addition, $\beta$ takes a larger value for each iteration.
Moreover, $\alpha$ serves as a distance threshold, and distances of less than $\alpha$ are considered sufficiently close.
If $d_{i, j}$ is small, $m_{i, j}$ takes a larger value and can be regarded as the similarity between $\mathbf{x}_i$ and $\mathbf{y}_j$.
Next, Sinkhorn's algorithm~\cite{sinkhorn} normalizes $\mathbf{M}$.
From the normalized $\mathbf{M}$, we obtain the correspondence between the points.

The orthogonal Procrustes method~\cite{procrustes} analytically calculates the optimal rigid transformation $T$ (iii).
To deal with a partial overlap, we adopted the same strategy as in \cite{rpmnet, dcp}.
We used $\sum_i^N{m_{i,j}}$ to weigh the correspondences.
A point with no correspondence has a weight of nearly 0.
Then, repeating the procedure from the feature extraction to applying a rigid transformation improves the accuracy (iv).
In RPMNet, a feature extraction is a bottleneck process.
Our experiments using an NVIDIA Tesla V100 confirmed that RPMNet takes approximately 80~ms to align approximately 1,000 objects, and feature extraction occupies more than 50\% of the computational time.

\section{Proposed method}
Our main idea is to reduce the computational cost by replacing PointNet with an extremely lightweight layer and a single linear layer with ReLU.
We call this single linear layer with ReLU a quasi-MLP (QMLP) because it is equivalent to an MLP when it is cascaded.
One may wonder if replacing PointNet with such a shallow layer would result in a loss of accuracy.
However, we found that it is possible to achieve the same level of accuracy by cascading the QMLPs during each iteration.
First, we describe the proposed cascade feature extraction method in Sec.~\ref{sec:method.faeture}, and in Sec.~\ref{sec:method.qmlp}, explain why QMLP has the same expressive power as an MLP.

\subsection{Cascading feature extraction}
\label{sec:method.faeture}
First, we describe the feature extraction in the first iteration.
The bottom part of Fig.~\ref{fig:overview} shows the configuration of our network.
Similar to the baseline, for feature extraction of a single point, the proposed method collects $K$-nearest neighbor points and calculates a classical feature vector $\mathbf{h}$ for each point.
However, unlike RPMNet, the proposed method does not include the absolute coordinate information in $\mathbf{h}$, and thus $\mathbf{h} \in \mathbb{R}^{7}$.
First, we input $\{\mathbf{h}_k\}_{k=1}^K$ individually into $\mathrm{MLP}: \mathbb{R}^{7} \to \mathbb{R}^{96}$ to obtain a 96-dimensional feature vector. 
\begin{eqnarray}
    \mathbf{g}_k = \mathrm{MLP}(\mathbf{h}_k).
\end{eqnarray}
We then take the maximum value of the obtained feature vectors for each dimension and obtain an intermediate feature vector $\mathbf{f}^{(0)} \in \mathbb{R}^{96}$.
\begin{eqnarray}
    \mathbf{f}^{(0)} = \mathrm{maxpool}(\mathcal{G}) \in \mathbb{R}^{96},
\end{eqnarray}
where $\mathcal{G} = \{\mathbf{g}_k\}_{k=1}^K \subset \mathbb{R}^{96}$.
This is the process for a single point.
By doing this for all points in the point cloud, we obtain a set of feature vectors $\mathcal{F}^{(0)} \subset \mathbb{R}^{96}$.
Note that the superscript $(i)$ denotes the number of iterations.

Next, we describe the feature extraction for the second and subsequent iterations.
In the $i + 1$ iteration of the proposed method, the set of feature vectors $\mathcal{F}^{(i)}$ and the absolute positions of points $\mathcal{X}^{(i)} \subset \mathbb{R}^{3}$ obtained in the $i$-th iteration are input into $\mathrm{QMLP}^{(i+1)}: \mathbb{R}^{96} \to \mathbb{R}^{96}$.
Let $\mathbf{x}^{(i)} \in \mathcal{X}^{(i)}$ be the absolute position of a point, and $\mathbf{f}^{(i)} \in \mathcal{F}^{(i)}$ be the corresponding set of feature vectors at the $i$-th iteration.
We obtain a feature vector $\mathbf{f}^{(i+1)}$ for the $i + 1$-th iteration as follows:
\begin{eqnarray}
    \mathbf{f}^{(i+1)} = \mathrm{QMLP}^{(i+1)} \left (\mathbf{f}^{(i)} + \mathbf{B}^{(i+1)}\mathbf{x}^{(i)} \right ).
\end{eqnarray}
Here, $\mathbf{B}^{(i+1)} \in \mathbb{R}^{96 \times 3}$ is a thin matrix, which will be discussed in Sec.~\ref{sec:method.qmlp}.

The critical differences between our design and that of RPMNet are two-fold.
(1) We use $\mathbf{f}^{(i)}$ to calculate $\mathbf{f}^{(i + 1)}$.
This design is the origin of our paper titled ``Cascading feature extraction''
(2) As detailed in Secs.~\ref{sec:method.faeture} and \ref{sec:method.qmlp}, the number of operations in our scheme is much smaller than that in RPMNet.

The proposed cascading feature extraction method utilizes the nature of simulated annealing.
Features are not always required to be fully discriminative at the beginning of the iterations because we apply only a coarse alignment.
Therefore, even if QMLP replaces several layered MLPs, the final accuracy often remains unchanged.
By contrast, at the end of the iterations, because the feature vectors passed through several QMLPs in the previous iterations, a single QMLP can conduct an equivalent computation by passing through the MLPs of the deeper layers.
Thus, QMLP, the computational cost of which is low, replaces PointNet.

\subsection{QMLP}
\label{sec:method.qmlp}

\begin{figure}[tb]
    \vspace{-3mm}
    \subfigure[]{
    \centering
    \hspace{-7mm}
    \includegraphics[height=31mm]{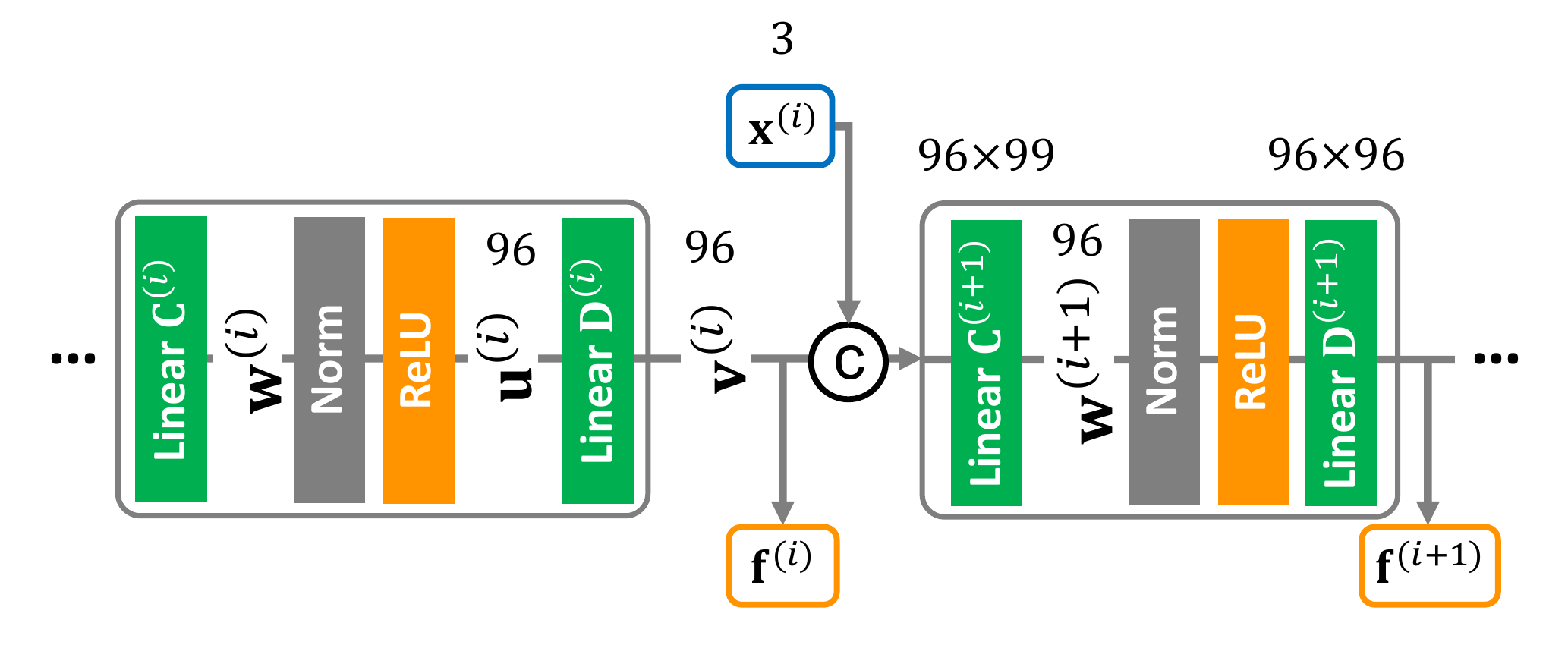} 
    \hspace{-4mm}
    \label{fig:layers}
    }
    \subfigure[]{
    \centering
    \includegraphics[height=31mm]{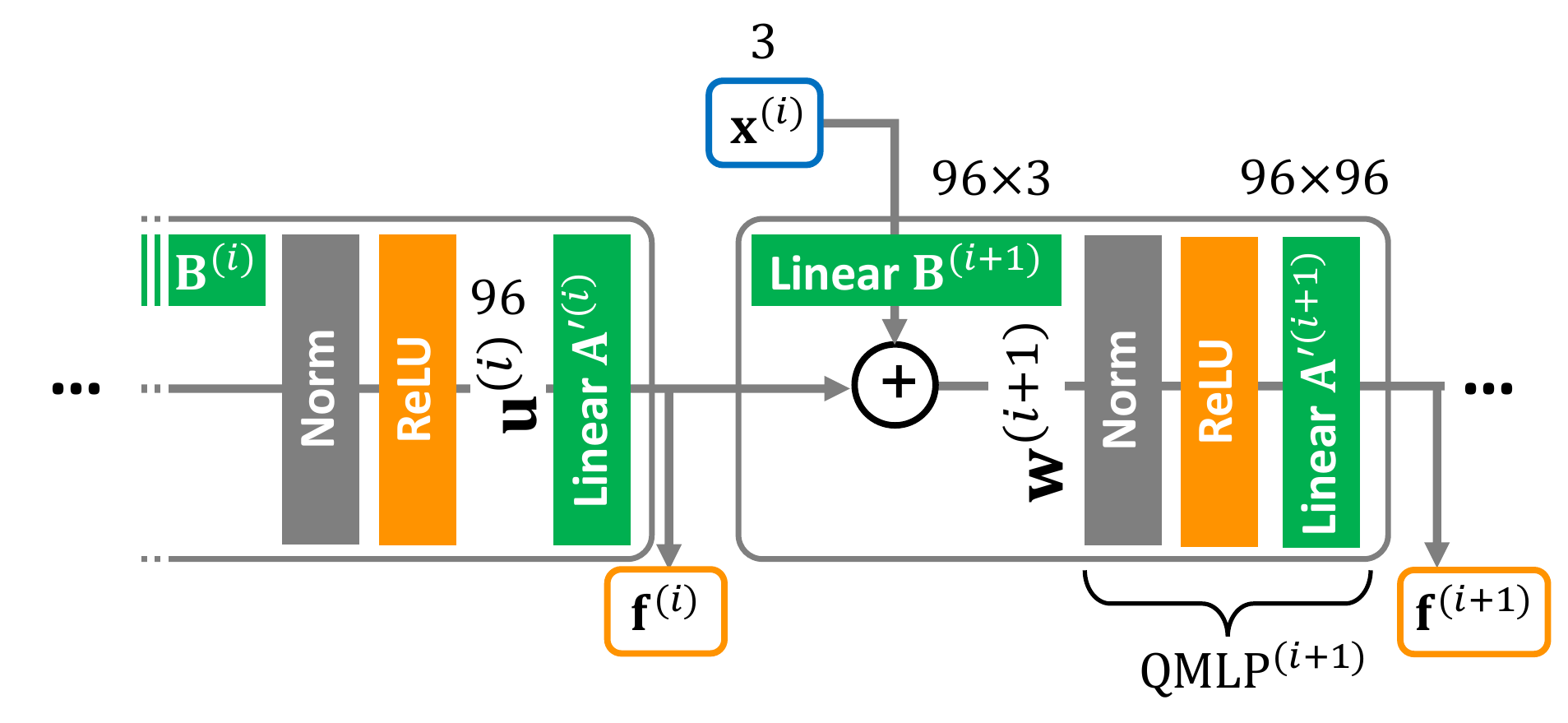} 
    \label{fig:layers_proposed}
    }
\vspace{-4mm}
\caption{(a)~Configuration of a cascaded feature extractor using an ordinary MLP (i.e., PointNet). The linear layers are successive between iterations. (b) Feature extractor constructed using the proposed QMLP. \textcircled{c} indicates a concatenation, and \textcircled{+} denotes an addition.}
\end{figure}

In terms of matrix operations, we show that cascading QMLPs are equivalent to the cascading ordinal MLPs used in PointNet and other applications.
QMLP consists of one linear transformation and one ReLU layer.
In the following, the matrix size and other values are the same as those in the actual implementation.
The MLP used in PointNet contains at least two linear transformations.
Each layer connects the input layer to the hidden layer and the hidden layer to the output layer.

We found that a cascaded feature extractor with an ordinary MLP is redundant.
Let us first model the ordinary MLP as follows.
Here, two linear transformations continue in succession between iterations, as shown in Fig.~\ref{fig:layers}.
This is a typical configuration of PointNet.
In Figs.~\ref{fig:layers} and \ref{fig:layers_proposed}, each block represents an MLP in the $i$-th iteration.
The input to the MLP passes through the first linear layer $\mathbf{C}^{(i)}\in\mathbb{R}^{96\times99}$, resulting in a vector $\mathbf{w}^{(i)}\in\mathbb{R}^{96}$. Next, normalization and ReLU layers are applied to $\mathbf{w}^{(i)}$ to obtain a hidden vector $\mathbf{u}^{(i)} \in \mathbb{R}^{96}$.
Subsequently, the second linear layer $\mathbf{D}^{(i)}\in\mathbb{R}^{96\times96}$ transforms $\mathbf{u}^{(i)}$ into the final output $\mathbf{v}^{(i)} \in \mathbb{R}^{96}$. 
The input to the MLP of the next iteration is a stack of $\mathbf{x}^{(i)}$ and $\mathbf{v}^{(i)}$.

Here, we show that we can reduce the number of calculations by rewriting the operation. Recall from Fig.~\ref{fig:layers} that we can obtain $\mathbf{w}^{(i+1)}$ from $\mathbf{C}^{(i+1)}$, $\mathbf{v}^{(i)}$, and $\mathbf{x}^{(i)}$. The equation can then be rewritten as follows:
\begin{eqnarray}
    \label{eq:mat_first}
    \mathbf{w}^{(i+1)} =& \mathbf{C}^{(i+1)}
    \left[ \begin{array}{c}
        \mathbf{v}^{(i)} \\
        \mathbf{x}^{(i)}
    \end{array}
    \right] \\
    =& 
    \left[ \begin{array}{cc}
        \mathbf{A}^{(i+1)} \mid \mathbf{B}^{(i+1)}
        \end{array}
    \right]
    \left[ \begin{array}{c}
        \mathbf{v}^{(i)} \\
        \mathbf{x}^{(i)}
    \end{array}
    \right] \\
    =& 
    \mathbf{A}^{(i+1)} \mathbf{v}^{(i)} + \mathbf{B}^{(i+1)} \mathbf{x}^{(i)} \\
    =& 
    \mathbf{A}^{(i+1)} \mathbf{D}^{(i)} \mathbf{u}^{(i)} + \mathbf{B}^{(i+1)} \mathbf{x}^{(i)} \\
    \label{eq:mat_last}
    =& 
    \mathbf{A'}^{(i)} \mathbf{u}^{(i)} + \mathbf{B}^{(i+1)} \mathbf{x}^{(i)}.
\end{eqnarray}
In this study, we decompose $\mathbf{C}^{(i+1)}\in\mathbb{R}^{96\times99}$ into smaller matrices $\mathbf{A}^{(i+1)}\in\mathbb{R}^{96\times96}$ and $\mathbf{B}^{(i+1)}\in\mathbb{R}^{96\times3}$.
We then define a new matrix $\mathbf{A'}^{(i)} = \mathbf{A}^{(i+1)} \mathbf{D}^{(i)} \in \mathbb{R}^{96\times96}$.
The final representation of Eq.~\ref{eq:mat_last} is our proposed module, which we visualize in Fig.~\ref{fig:layers_proposed}.
Equation~\ref{eq:mat_last} implies that if we smartly configure the matrices, we can compute $\mathbf{w}^{(i+1)}$ simply by (1) the output of the previous step ($\mathbf{A'}^{(i)}\mathbf{u}^{(i)}$) and (2) the current point position $\mathbf{x}^{(i)}$ with a shallow transformation $\mathbf{B}^{(i+1)}$.

In a conventional MLP, two matrix operations using matrices $\mathbf{C}$ and $\mathbf{D}$ are conducted on the input vector (the green modules in Fig~\ref{fig:layers}).
By contrast, with the proposed method, the costly operation uses only $\mathbf{A}^{'}$ (the green modules in Fig~\ref{fig:layers_proposed}).
This is because the size of $\mathbf{B}$ is much smaller than that of $\mathbf{A}^{'}$.

These results show that the proposed cascaded feature extractor can achieve the same performance as an ordinary MLP while reducing the computational cost through the use of QMLP.

\subsection{Analysis}
\begin{figure}[bt]
    \hspace{-8pt}
    \includegraphics[width=1.04\hsize]{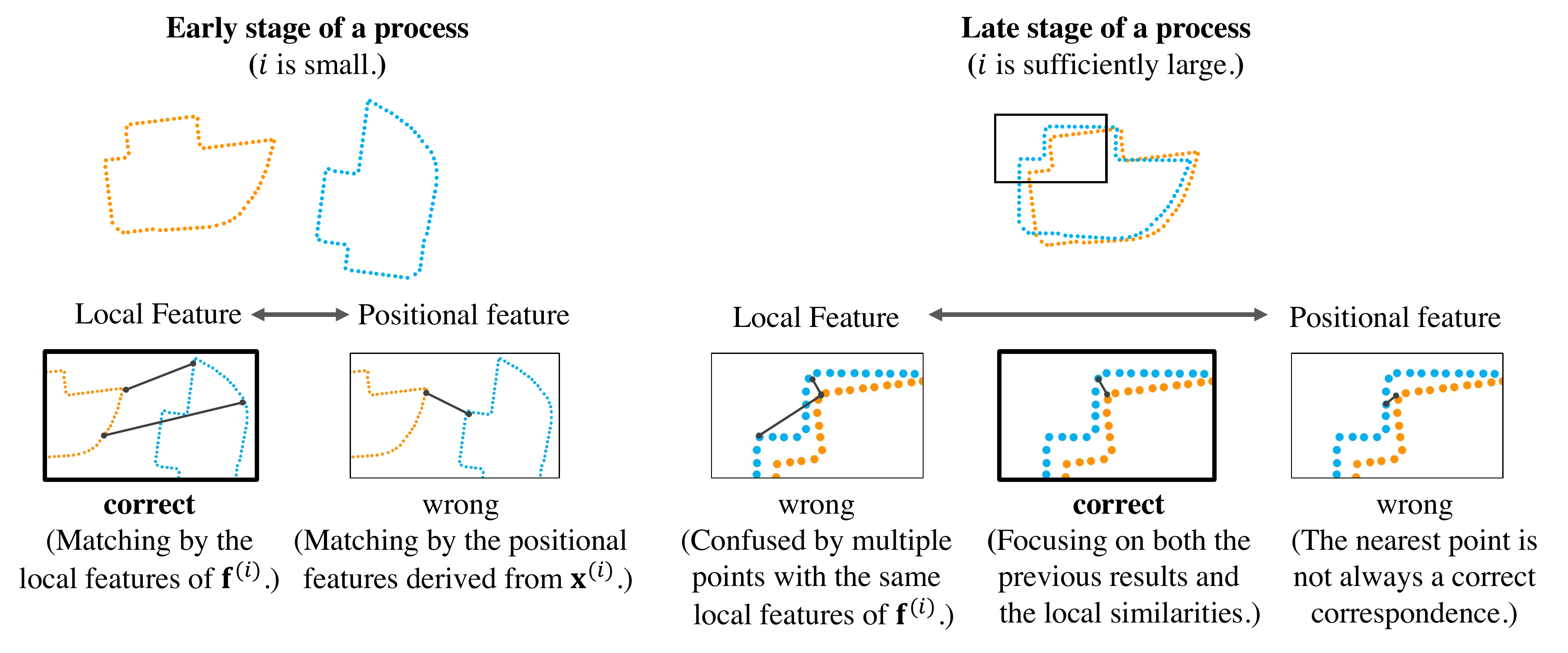}
    \vspace{-5mm}
    \caption{
        Change in pattern to be identified between the beginning and end of the iteration.
        At the end of the iteration, only the close points in the absolute position are in correspondence.
        }
    \label{fig:stagewise}
\end{figure}

\paragraph{Accurate Correspondence}
Next, we show why the proposed method does not degrade the accuracy, it occasionally improves it.
The patterns that we should focus on differ depending on the registration stage.
First, at the beginning of the iterations, QMLP needs to focus on the local feature of $\mathbf{f}^{(i)}$ rather than the positional feature derived from $\mathbf{x}^{(i)}$.
This is because the information of the absolute position does not help find the correspondence during this stage.
By contrast, at the end of the iterations, focusing on both the local and positional features leads to a correct correspondence.
Fig.~\ref{fig:stagewise} illustrates this.
There are many points whose local features are similar.
However, because we roughly aligned the point clouds at the beginning of the iterations, the only correct counterparts are already close in position.

The baseline used PointNet, which has the same weight parameters for all iterations.
This means that it focuses on the same pattern during all stages of alignment.
By contrast, with our proposed method, the feature extractor parameters used in each iteration are all different.
Therefore, our proposed method can adaptively focus on different patterns at different stages of alignment.

\vspace{-3mm}
\paragraph{Computational Cost}
The baseline uses $K$-nearest neighbors for the feature extraction of a single point.
It applies the same feature extraction every iteration, and thus has a computational complexity of $\mathcal{O}(D^{2}KL)$.
The dimensionality of the feature vectors is $D$.
The number of neighbors used for feature extraction is $K$.
The number of iterations is $L$.
Herein, we approximate the running time of PointNet, where the maximum width of $D$ is $\mathcal{O}(D^2)$.
By contrast, the proposed method has a computational complexity of $\mathcal{O}(KD^{2})$ in the first iteration, as in PointNet, and $\mathcal{O}(D^2)$ in the second and subsequent iterations, and thus the total computational complexity is $\mathcal{O}(D^2(K + L))$.
This is indeed smaller than $\mathcal{O}(D^2KL)$.

\subsection{Other Ingenuities for Speedup}
We propose two other ingenuities to accelerate the method: batch annealing parameter extraction and Sinkhorn normalization.

\vspace{-3mm}
\paragraph{Batch Prediction of Annealing Parameters} A network separate from the feature extraction module predicts the appropriate parameters ($\alpha$ and $\beta$) at each iteration in the baseline.
Thus, the cost of annealing parameter prediction increases according to the number of iterations.
However, with the proposed method, the parameters for each iteration are estimated simultaneously in the first iteration.
Therefore, we were also able to reduce the computational complexity of the parameter estimation for use in the annealing.
This reduced the computational cost of the annealing parameter prediction by a factor of five.

\vspace{-3mm}
\paragraph{Adaptive Iteration for Sinkhorn Normalization} The number of iterations of Sinkhorn normalization is constant in the baseline.
By contrast, we changed the number of iterations depending on the alignment stage to speed up the process.
Initially, we reduce the number of iterations,
and then increase the number to a sufficient value.
In our implementation, we empirically set the number to equal the number of current registration iterations (i.e., if $i = 4$, the number of iterations of Sinkhorn normalization is also set to four).
This operation accelerates Sinkhorn normalization 1.5-fold.
This is to take advantage of the fact that the early stages of alignment do not require precision.

\subsection{Code-Level Optimization}
\label{sec:method.imple}
We found that the original implementation of RPM-Net has room to accelerate at the implementation level.
Thus, we improved RPM-Net~\cite{rpmnet} for three points: Sinkhorn normalization, singular value decomposition (SVD), and a nearest neighbor search.
We call the result ``RPM-Net (Optim)'', which is 20 ms faster than the original implementation (see Fig.~\ref{fig:acc_time}).
We use RPM-Net (Optim) for our comparison, which is the basis of our
proposed method.

\vspace{-3mm}
\paragraph{Sinkhorn Normalization} We accelerated the runtime of Sinkhorn normalization by three-fold through the following modification.
As the baseline, the authors integrated Sinkhorn normalization and the calculation of the similarities using the exponential, following the method of Mena~\etal~\cite{latent_permutations}.
The implementation is shown in Algorithm~\ref{alg:sk_log}.
However, the computation costs of the logarithmic and exponential functions are much higher than those of addition and division.
Therefore, we reimplemented the Sinkhorn normalization in Algorithm \ref{alg:sk_sum}, where $l$ is the number of iterations of the Sinkhorn normalization.

\vspace{-3mm}
\paragraph{SVD} We accelerated the SVD by computing it on a CPU instead of a GPU.
In the existing code, all registration algorithms, including SVD, were executed on a GPU.
We profiled the execution time of the SVD on a GPU and found that it took approximately 2~ms to process a $3 \times 3$ matrix.
We studied this process in detail and found that SVD is faster on a CPU even after including the transfer cost to the CPU.

The reason for this may be the difference in performance between the GPU and CPU on a single core.
If a problem is small and difficult to parallelize, GPUs are slower than CPUs.
Here, the size of the covariance matrix of the point cloud handled by the orthogonal Procrustes method is $3 \times3$, which cannot be parallelized.
Because the matrix to be transferred is extremely small, the communication time between the CPU and GPU is negligible.
With this revision, the processing time of the Orthogonal Procrustes algorithm was halved.

\vspace{-3mm}
\paragraph{Nearest Neighbor Search} We used Faiss~\cite{faiss} to search for the $K$-nearest neighbors for the feature extraction of each point.
Faiss is a fast library, particularly for large-scale data, and we therefore applied it during the experiment using KITTI~\cite{kitti}.
When the number of points is 12,000, the $K$-nearest neighbor search becomes six times faster than the top-k function of PyTorch~\cite{pytorch}.

\vspace{-6mm}
\hspace{-6mm}
\begin{minipage}[t]{0.53\textwidth}
\begin{algorithm}[H]
\caption{Implementation by Mena \etal{}~\cite{latent_permutations}\\} 
\begin{algorithmic}[1]
    \label{alg:sk_log}      
        \WHILE {$c < l$}
            \STATE $d_{i,j} \leftarrow d_{i,j} - \mathrm{log}(\sum_i{e^{d_{i,j}}}), \forall i, j$
            \STATE $d_{i,j} \leftarrow d_{i,j} - \mathrm{log}(\sum_j{e^{d_{i,j}}}), \forall i, j$
            \STATE $c \leftarrow c + 1$
        \ENDWHILE
        \STATE $m_{i,j} \leftarrow \exp{({d_{i,j}}^2)}, \forall i, j$
\end{algorithmic}
\end{algorithm}
\end{minipage}
\hfill
\begin{minipage}[t]{0.43\textwidth}
\begin{algorithm}[H]
\caption{Standard implementation of Sinkhorn normalization}         
\begin{algorithmic}[1]
    \label{alg:sk_sum}                          
        \STATE $m_{i,j} \leftarrow \exp{({d_{i,j}}^2)}, \forall i, j$
        \WHILE {$c < l$}
        \STATE $m_{i,j} \leftarrow m_{i,j} / \sum_i{m_{i,j}}, \forall i, j$
        \STATE $m_{i,j} \leftarrow m_{i,j} / \sum_j{m_{i,j}}, \forall i, j$
        \STATE $c \leftarrow c + 1$
        \ENDWHILE
\end{algorithmic}
\end{algorithm}
\end{minipage}

\section{Experiment}
\vspace{-2mm}
\subsection{Environment}
\vspace{-2mm}
We used the GPU cluster of the National Institute of Advanced Industrial Science and Technology, AI Bridging Cloud Infrastructure (ABCI), to measure the speed of the model.
The GPU is an NVIDIA Tesla V100, the CPU is an Intel Xeon Gold 6148, and the main memory is 32-GB DDR4 2666-MHz RDIMM (ECC).

\vspace{-2mm}
\subsection{Datasets}
\vspace{-2mm}
The datasets used in our experiments are modelnet40~\cite{modelnet40} and KITTI~\cite{kitti}.
Here, modelnet40 is a 40-class 3D point cloud object.
We preprocessed modelnet40 in a similar manner as the methods developed by Yew and Lee~\cite{rpmnet} and Wang and Solomon~\cite{dcp}.
We used approximately 5,100 point cloud objects from the first 20 classes for training, approximately 1,200 for validation, and approximately 1,300 from the second 20 classes for the test.
The number of points in each object was 2,048, and 1,024 points were randomly selected for the test.

For the input data, approximately 70\% of the point clouds were selected from a single point cloud object, and the point clouds to be aligned have an overlap of approximately 40\%.
Furthermore, Gaussian noise was applied to each point position of the objects, and one of the objects was rotated and translated.
The rotations are within the range [0, 45\textdegree] for three Euler angles.
Translations are within the range [-0.5, 0.5] in modelnet40, and [-15, 15] in KITTI.

\begin{figure}[tb]
    \center
    \vspace{-1mm}
    \includegraphics[keepaspectratio, width=\hsize]{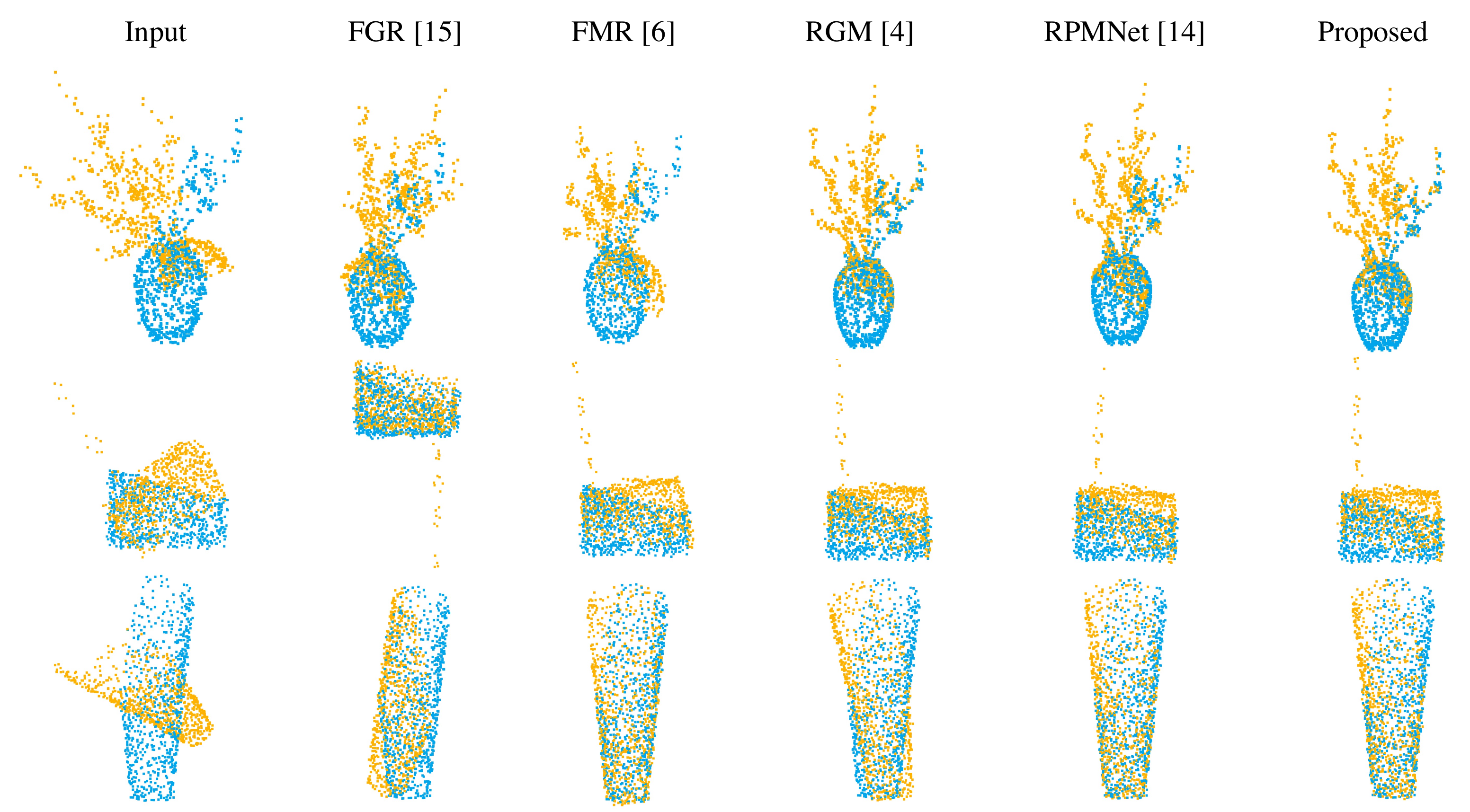}
    \vspace{-7mm}
    \caption{Qualitative comparison}
    \vspace{0mm}
    \label{fig:modelnet40_examples}
\end{figure}
KITTI is made up of point cloud data from an urban area.
KITTI point cloud data come with GPS location information.
Based on the location information, we can create the teacher data for positioning.
However, owing to the large error in GPS positioning, we follow the method of Choy \etal~\cite{fcgf} and refine the data using the ICP algorithm. During each stage, one of the point clouds was rotated randomly to the initial position.
The model trained by modelnet40 was retrained using KITTI and evaluated.

\vspace{-2mm}
\subsection{Models}
\vspace{-2mm}
In this section, we describe the test parameters of the proposed and the comparison methods.
There are 64 $K$-nearest neighbor points for feature extraction and five registration iterations.

To make a fair comparison, we first conducted a code-level acceleration on the existing RPMNet, which we
call RPM-Net (Optim).
Although this achieves the same accuracy as RPMNet, it is 40\% to 200\% faster.
This speedup includes the optimization described in Sec.~\ref{sec:method.imple}.

We compared our approach with ICP~\cite{icp}, FGR~\cite{fgr}, DCP~\cite{dcp}, FMR~\cite{fmr}, and RGM~\cite{Fu2021RGM}.
Among these methods, ICP and FGR are classical methods that are not learning-based and are computed on a CPU.
The other methods are based on deep learning and are executed on a GPU.
In addition, FMR does not use a feature matching scheme, and
RGM is the latest method to apply graph matching.
\vspace{-2mm}
\subsection{Results and Evaluation}

\begin{wraptable}[12]{r}[2mm]{0.66\hsize}
    \vspace{-4mm}
    \centering
    \hspace{2mm}
    \begin{tabular}{@{}l@{}rrlr@{}} \toprule
    method & RE $\downarrow$ & TE $\downarrow$ & CD $\downarrow$ & time $\downarrow$ \\ \midrule
    ICP~\cite{icp}~(CPU) & 27.25 \textdegree & 28 cm & 0.015 & 11 ms\\
    FGR~\cite{fgr}~(CPU) & 31.43 \textdegree & 20 cm & 0.012 & 34 ms\\
    DCP-V2~\cite{dcp} & 12.61 \textdegree & 17 cm & 0.011 & 12 ms\\ 
    FMR~\cite{fmr} & 12.14 \textdegree & 17 cm & 0.0086 & 54 ms \\
    RPMNet~(Optim)~\cite{rpmnet} & 1.71 \textdegree & 1.8 cm & 0.00085 & 58 ms\\
    RGM~\cite{Fu2021RGM} & 1.56 \textdegree & 1.5 cm & 0.00084 & 174 ms\\
    Proposed & 1.23 \textdegree & 1.3 cm & 0.00076 & 17 ms\\ \bottomrule
    \end{tabular}
    \vspace{-2mm}
    \caption{Speed and accuracy table for modelnet40. The proposed method is more accurate than the baseline.}
    \label{tb:compare_baseline}
\end{wraptable}
For the above dataset, we conducted a point cloud registration and measured its accuracy and speed.
Table~\ref{tb:compare_baseline} and Fig.~\ref{fig:acc_time} show the results.
We used the modified chamfer error introduced in ~\cite{rpmnet} (``CD'' in Table~\ref{tb:compare_baseline}).
The accuracies of RPMNet, RGM, and our proposed method are much higher than those of the other methods.
We can confirm that our proposed method is faster than the other models with high accuracy.

\begin{wraptable}[9]{r}[0mm]{0.59\hsize}
    \centering
    \vspace{-3mm}
    \begin{tabular}{@{}l@{}rr@{}r@{}r@{}} \toprule
    method & RE $\downarrow$ & TE $\downarrow$ & \shortstack{4,000 pts\\ time}$\downarrow$ & \shortstack{12,000 pts\\ time}$\downarrow$\\ \midrule
    RPMNet~(Optim) & - & - & 200 ms & 750 ms\\
    FGR & 5.52 \textdegree & 3.53 m & 250 ms & 1600 ms\\
    FMR & 1.89 \textdegree & 1.18 m & 67 ms & 96 ms \\
    Proposed & 1.66 \textdegree & 0.92 m & 56 ms & 220 ms\\ \bottomrule
    \end{tabular}
    \vspace{-2mm}
    \caption{Evaluation on KITTI dataset}
    \label{tb:kitti}
\end{wraptable}
Table~\ref{tb:compare_baseline} also shows that the proposed method outperforms the baseline and RGM in all evaluation metrics, i.e., rotation error, translation error, and chamfer distance.
This can be attributed to the fact that the proposed method extracts features with different weight parameters for each iteration and can adaptively focus on different patterns depending on the alignment stage.

\begin{wraptable}[7]{r}[2mm]{0.35\hsize}
    \centering
    \vspace{-4mm}
    \begin{tabular}{@{}ll@{}} \toprule
    method & CD $\downarrow$ \\ \midrule
    Proposed w/ Sec~\ref{sec:method.faeture} & 0.00076 \\
    Proposed w/o Sec~\ref{sec:method.faeture} & 0.0015 \\ \bottomrule
    \end{tabular}
    \vspace{-2mm}
    \caption{Ablation study of cascading feature extraction.}
    \label{tb:fix_weight}
\end{wraptable}
Fig.~\ref{fig:modelnet40_examples} shows a qualitative comparison. 
Our proposed method and RPMNet achieved similar results, whereas our method is more accurate in terms of the details (e.g., branch of a tree in the first row) and three times faster.
Both qualitative and quantitative comparisons show that our method is more accurate than RGM.

Table~\ref{tb:fix_weight} shows an ablation study of the accuracy,
using (1) separate weight parameters for each iteration and (2) the same weight parameters for all iterations.
The result of ``Proposed w/o Sec~\ref{sec:method.faeture}'' is less accurate than that of ``Proposed w/ Sec~\ref{sec:method.faeture}.''
This result justifies the explanation of Sec~\ref{sec:method.faeture}.

\begin{wrapfigure}[17]{r}[3mm]{0.46\hsize}
    \vspace{-5mm}
    \includegraphics[width=\hsize]{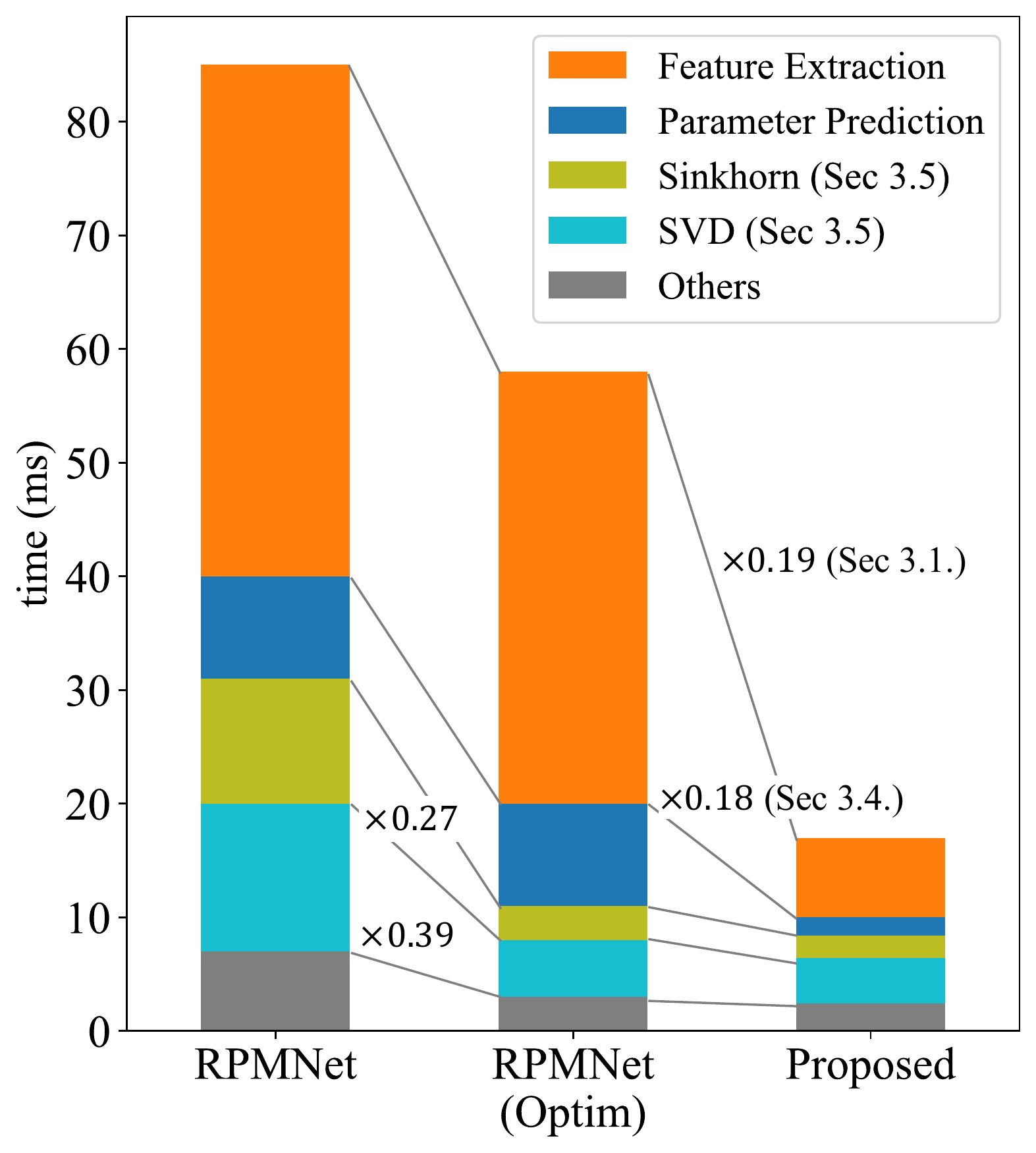}  
    \vspace{-8mm}
    \caption{Module-wise speed comparison.}
    \label{fig:ablation}
\end{wrapfigure}

Fig.~\ref{fig:ablation} shows a comparison of the speed for each module. Code-optimization mainly improves the speed of the SVD and Sinkhorn normalization.
Our proposed cascading feature extraction significantly speeds up the feature extraction, which has been a bottleneck.

\begin{wrapfigure}[7]{r}[5mm]{0.58\hsize}
    \vspace{-17mm}
    \includegraphics[width=\hsize]{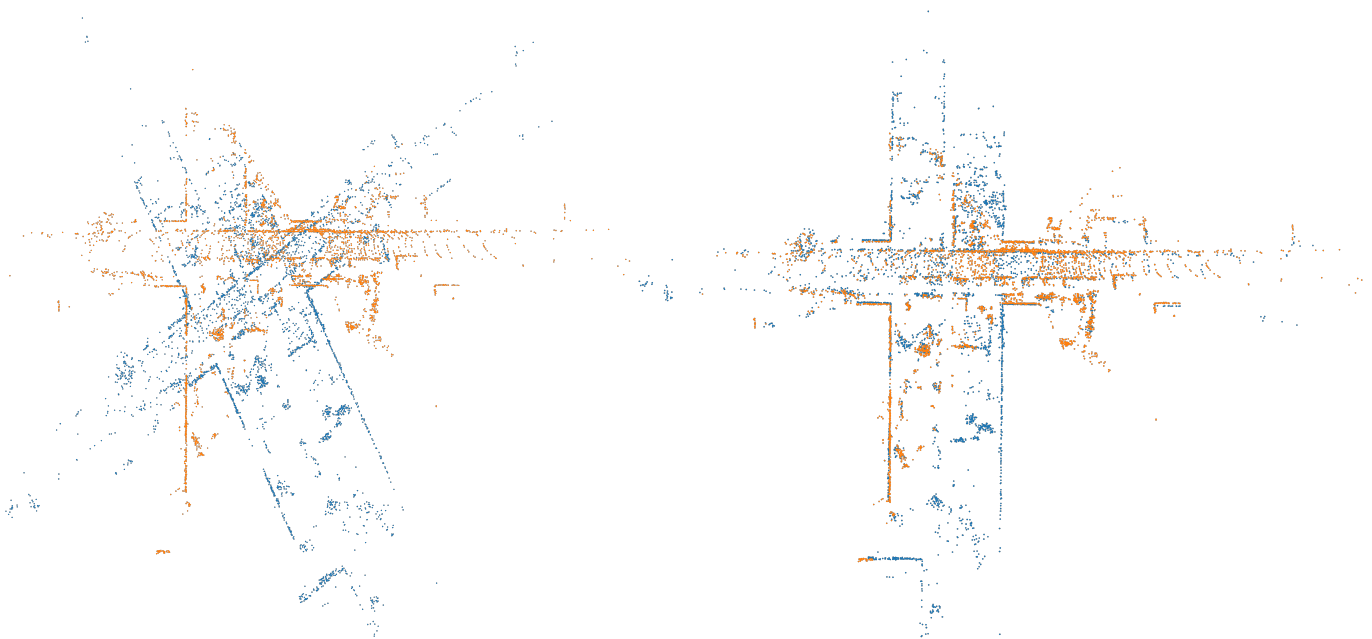} 
    \vspace{-8mm}
    \caption{Example registration result of KITTI}  
    \label{fig:kitti_example}
\end{wrapfigure}

Table~\ref{tb:kitti} shows the speed and accuracy of the experiments using KITTI.
Fig.~\ref{fig:kitti_example} shows the results of the actual registration.
We can confirm that the registration is accurate.
The size of the point cloud used for the accuracy evaluation is 4,000.
The proposed method is three times faster than the baseline, even for large point clouds such as KITTI.

Because the memory consumption of the baseline, RPMNet, is extremely large and difficult to train, we did not evaluate its accuracy on a large point cloud but only measured its inference time.
Notably, FMR is the fastest for 12,000 points.
FMR aligns two point clouds such that the feature vectors of the points are close.
Specifically, it operates as follows:
(1) The point cloud is perturbated; 
(2) A transformation is found and applied to the point cloud to minimize the error between the feature vectors of the point clouds;
(3) The point cloud is perturbated again and the process is repeated until convergence is achieved.
Thus, FMR does not explicitly depend on the number of points.
This speeds up the FMR for a massive number of points.
However, the overheads of (1) and (2) are significant even if the number of points is small.
This overhead makes FMR relatively slower for small- and middle-scale data.

\subsection{Limitation}
Our method has two limitations.
As the first condition, because we prepared individual weight parameters for each iteration, we need to specify the number of alignment iterations during the training phase.
The baseline RPMNet does not have this limitation.

The second edition is scalability.
Similar to other existing methods such as RPMNet~\cite{rpmnet} and RGM~\cite{Fu2021RGM}, our method also uses Sinkhorn normalization, which has a computational complexity of the square of the number of points.
Thus, it becomes a bottleneck when we increase the number of points, as with other methods.

\vspace{-3mm}
\section{Conclusion}
We proposed a method for speeding up an iterative 3D point cloud registration.
As our main idea, we do not have to compute the feature vectors until they are fully discriminative.
We showed that our cascading feature extraction is computationally efficient.
The inference speed is three times faster than that of the baseline.
Furthermore, the accuracy is higher than that of the latest models, such as RPMNet and RGM.

\section{Acknowledgements}
This work was supported by JST, PRESTO, Japan, Grant Number JPMJPR 1936.

\bibliography{main}

\begin{thebibliography}{15}
\providecommand{\natexlab}[1]{#1}
\providecommand{\url}[1]{\texttt{#1}}
\expandafter\ifx\csname urlstyle\endcsname\relax
  \providecommand{\doi}[1]{doi: #1}\else
  \providecommand{\doi}{doi: \begingroup \urlstyle{rm}\Url}\fi

\bibitem[{Arun} et~al.(1987){Arun}, {Huang}, and {Blostein}]{icp}
K.~S. {Arun}, T.~S. {Huang}, and S.~D. {Blostein}.
\newblock Least-squares fitting of two 3-d point sets.
\newblock volume PAMI-9, pages 698--700, 1987.
\newblock \doi{10.1109/TPAMI.1987.4767965}.

\bibitem[{Charles} et~al.(2017){Charles}, {Su}, {Kaichun}, and
  {Guibas}]{pointnet}
R.~Q. {Charles}, H.~{Su}, M.~{Kaichun}, and L.~J. {Guibas}.
\newblock Pointnet: Deep learning on point sets for 3d classification and
  segmentation.
\newblock In \emph{2017 IEEE Conference on Computer Vision and Pattern
  Recognition (CVPR)}, pages 77--85, 2017.

\bibitem[Choy et~al.(2019)Choy, Park, and Koltun]{fcgf}
Christopher Choy, Jaesik Park, and Vladlen Koltun.
\newblock Fully convolutional geometric features.
\newblock In \emph{Proceedings of the IEEE International Conference on Computer
  Vision (CVPR)}, pages 8958--8966, 2019.

\bibitem[Fu et~al.(2021)Fu, Liu, Luo, and Wang]{Fu2021RGM}
Kexue Fu, Shaolei Liu, Xiaoyuan Luo, and Manning Wang.
\newblock Robust point cloud registration framework based on deep graph
  matching.
\newblock \emph{Internaltional Conference on Computer Vision and Pattern
  Recogintion (CVPR)}, 2021.

\bibitem[Geiger et~al.(2013)Geiger, Lenz, Stiller, and Urtasun]{kitti}
Andreas Geiger, Philip Lenz, Christoph Stiller, and Raquel Urtasun.
\newblock Vision meets robotics: The kitti dataset.
\newblock \emph{International Journal of Robotics Research (IJRR)}, 2013.

\bibitem[Huang et~al.(2020)Huang, Mei, and Zhang]{fmr}
Xiaoshui Huang, Guofeng Mei, and Jian Zhang.
\newblock Feature-metric registration: A fast semi-supervised approach for
  robust point cloud registration without correspondences.
\newblock In \emph{The IEEE/CVF Conference on Computer Vision and Pattern
  Recognition (CVPR)}, 2020.

\bibitem[Johnson et~al.(2019)Johnson, Douze, and Jégou]{faiss}
Jeff Johnson, Matthijs Douze, and Hervé Jégou.
\newblock Billion-scale similarity search with gpus.
\newblock \emph{IEEE Transactions on Big Data}, pages 1--1, 2019.
\newblock \doi{10.1109/TBDATA.2019.2921572}.

\bibitem[Mena et~al.(2018)Mena, Belanger, Linderman, and
  Snoek]{latent_permutations}
Gonzalo Mena, David Belanger, Scott Linderman, and Jasper Snoek.
\newblock Learning latent permutations with gumbel-sinkhorn networks.
\newblock In \emph{International Conference on Learning Representations
  (ICLR)}, 2018.

\bibitem[Paszke et~al.(2019)Paszke, Gross, Massa, Lerer, Bradbury, Chanan,
  Killeen, Lin, Gimelshein, Antiga, Desmaison, Kopf, Yang, DeVito, Raison,
  Tejani, Chilamkurthy, Steiner, Fang, Bai, and Chintala]{pytorch}
Adam Paszke, Sam Gross, Francisco Massa, Adam Lerer, James Bradbury, Gregory
  Chanan, Trevor Killeen, Zeming Lin, Natalia Gimelshein, Luca Antiga, Alban
  Desmaison, Andreas Kopf, Edward Yang, Zachary DeVito, Martin Raison, Alykhan
  Tejani, Sasank Chilamkurthy, Benoit Steiner, Lu~Fang, Junjie Bai, and Soumith
  Chintala.
\newblock Pytorch: An imperative style, high-performance deep learning library.
\newblock In H.~Wallach, H.~Larochelle, A.~Beygelzimer, F.~d\textquotesingle
  Alch\'{e}-Buc, E.~Fox, and R.~Garnett, editors, \emph{Advances in Neural
  Information Processing Systems 32}, pages 8024--8035. Curran Associates,
  Inc., 2019.

\bibitem[Schönemann(1966)]{procrustes}
Peter Schönemann.
\newblock A generalized solution of the orthogonal procrustes problem.
\newblock \emph{Psychometrika}, 31\penalty0 (1):\penalty0 1--10, 1966.

\bibitem[Sinkhorn(1964)]{sinkhorn}
Richard Sinkhorn.
\newblock A relationship between arbitrary positive matrices and doubly
  stochastic matrices.
\newblock \emph{Ann. Math. Statist.}, 35\penalty0 (2):\penalty0 876--879, 1964.
\newblock \doi{10.1214/aoms/1177703591}.

\bibitem[Wang and Solomon(2019)]{dcp}
Yue Wang and Justin~M. Solomon.
\newblock Deep closest point: Learning representations for point cloud
  registration.
\newblock In \emph{The IEEE International Conference on Computer Vision
  (ICCV)}, 2019.

\bibitem[Wu et~al.(2015)Wu, Song, Khosla, Yu, Zhang, Tang, and
  Xiao]{modelnet40}
Zhirong Wu, Shuran Song, Aditya Khosla, Fisher Yu, Linguang Zhang, Xiaoou Tang,
  and Jianxiong Xiao.
\newblock 3d shapenets: A deep representation for volumetric shapes.
\newblock In \emph{The IEEE Conference on Computer Vision and Pattern
  Recognition (CVPR)}, pages 1912--1920. IEEE Computer Society, 2015.
\newblock ISBN 978-1-4673-6964-0.

\bibitem[Yew and Lee(2020)]{rpmnet}
Zi~Jian Yew and Gim~Hee Lee.
\newblock Rpm-net: Robust point matching using learned features.
\newblock In \emph{Conference on Computer Vision and Pattern Recognition
  (CVPR)}, 2020.

\bibitem[Zhou et~al.(2016)Zhou, Park, and Koltun]{fgr}
Qian-Yi Zhou, Jaesik Park, and Vladlen Koltun.
\newblock Fast global registration.
\newblock In \emph{2016 European Conference on Computer Vision (ECCV)}, pages
  766--782, 2016.

\end{thebibliography}
\end{document}